\documentclass{article}

\usepackage{arxiv}

\usepackage[utf8]{inputenc}
\usepackage[T1]{fontenc}
\usepackage{hyperref}
\usepackage{url}
\usepackage{booktabs}
\usepackage{amsfonts}
\usepackage{amsmath}
\usepackage{nicefrac}
\usepackage{microtype}
\usepackage{cleveref}
\usepackage{graphicx}
\usepackage{natbib}
\usepackage{doi}
\usepackage{float}

\title{Data-Driven Autoregressive Power Prediction for GTernal Robots in the Robotarium}

\newif\ifuniqueAffiliation
\uniqueAffiliationtrue

\ifuniqueAffiliation
\author{
    Yassin Abdelmeguid \\
    American University of Sharjah  \\
    \texttt{b00083915@aus.edu} \\
    \And
    Ammar Hasan \\
    American University of Sharjah  \\
    \texttt{amhasan@aus.edu} \\
}
\fi

\renewcommand{\shorttitle}{Autoregressive Power Prediction for GTernal Robots}

\hypersetup{
    pdftitle={Data-Driven Power Prediction for GTernal Mobile Robots},
    pdfsubject={cs.RO, cs.LG},
    pdfauthor={Author 1, Author 2, Author 3},
    pdfkeywords={mobile robotics, energy prediction, multi-robot systems, Robotarium, power consumption modeling},
}

\begin{document}
\maketitle

\begin{abstract}
Energy-aware algorithms for multi-robot systems require accurate power consumption models, yet existing approaches rely on kinematic approximations that fail to capture the complex dynamics of real hardware. We present a lightweight autoregressive predictor for the GTernal mobile robot platform deployed in the Georgia Tech Robotarium. Through analysis of 48,000 samples collected across six motion trials, we discover that power consumption exhibits strong temporal autocorrelation ($\rho_1 = 0.95$) that dominates kinematic effects. A 7,041-parameter multi-layer perceptron (MLP) achieves $R^2 = 0.90$ on held-out motion patterns by conditioning on recent power history, reaching the theoretical prediction ceiling imposed by measurement noise. Physical validation across seven robots in a collision avoidance scenario yields mean $R^2 = 0.87$, demonstrating zero-shot transfer to unseen robots and behaviors. The predictor runs in 224 $\mu$s per inference, enabling real-time deployment at 150$\times$ the platform's 30 Hz control rate. We release the trained model and dataset to support energy-aware multi-robot algorithm development.
\end{abstract}


\section{Introduction}

Multi-robot coordination algorithms increasingly incorporate energy awareness as a first-class objective, whether for coverage control with battery constraints, task allocation under power budgets, or reinforcement learning with energy-shaped rewards. These algorithms require power consumption models that can predict energy expenditure from commanded actions, enabling planners and policies to reason about battery depletion before committing to trajectories. Simulators for multi-robot systems similarly need accurate energy models to train policies that transfer to physical hardware without experiencing unexpected battery failures.

The dominant approach to energy modeling assumes power consumption follows directly from robot kinematics, typically through polynomial functions of velocity magnitude or motor commands \citep{hou2019energy}. While intuitive, this formulation ignores the complex thermal dynamics of motors, transient behaviors during acceleration, wireless communication overhead, and other hardware effects that contribute significantly to instantaneous power draw \citep{gora2024energy}. These unmodeled dynamics manifest as large prediction errors when kinematic models encounter real telemetry data.

We present a data-driven energy predictor for the GTernal mobile robot platform \citep{kim2026gternal} deployed in the Georgia Tech Robotarium \citep{wilson2020robotarium}. The Robotarium provides remote access to a swarm of differential-drive robots equipped with INA260 power monitors capable of measuring voltage, current, and power at 10 mW precision. This infrastructure enables systematic data collection across diverse motion patterns while capturing ground-truth power consumption at the 30 Hz command rate.

Our key insight emerges from statistical analysis of the collected data. Power consumption exhibits remarkably strong temporal autocorrelation, with lag-1 correlation $\rho_1 = 0.95$ across all motion trials. This autocorrelation structure implies that recent power history contains far more predictive information than current kinematic state, consistent with first-order autoregressive (AR(1)) process dynamics where future values depend linearly on immediate past observations \citep{triebe2019arnet}. A model that observes the previous power reading can explain 90\% of variance in the next reading, while velocity features alone explain only 16\%. The dominance of autoregressive structure over kinematic features suggests that power dynamics are governed by latent hardware states including motor thermal effects, WiFi transients, and sensor duty cycles that persist across timesteps but remain invisible to kinematic models.

We contribute the following. First, we characterize power consumption dynamics on physical GTernal robots through systematic data collection across six motion protocols totaling 26.5 minutes of operation. Second, we design and validate an autoregressive Multi-layer Perceptron (MLP) predictor that achieves $R^2 = 0.90$ at the theoretical noise ceiling using only 7,041 parameters. Third, we demonstrate zero-shot transfer to seven unseen robots executing collision avoidance behaviors, achieving mean $R^2 = 0.87$ without retraining. Fourth, we release the trained model, dataset, and deployment code at [https://github.com/y4ssn/robotarium-energy-predictor].

\section{Related Work}

Energy modeling for mobile robots has traditionally relied on physics-based formulations that decompose power consumption into kinetic, frictional, and resistive components \citep{hou2019energy}. However, comprehensive reviews of energy prediction techniques highlight that such analytical models struggle to account for computational overhead, sensor duty cycles, and thermal dynamics inherent to autonomous platforms \citep{gora2024energy}. Recent work has shifted toward data-driven approaches, with neural architectures demonstrating superior accuracy by implicitly learning unmodeled dynamics from telemetry \citep{visca2022probabilistic}.

Autoregressive neural networks have proven effective for energy prediction in related domains. Non-linear autoregressive models with exogenous inputs (NARX) successfully forecast building energy consumption by capturing thermal inertia alongside external driving variables \citep{ruiz2016narx}. Similar architectures have been applied to unmanned aerial vehicle flight dynamics, where data-driven temporal models circumvent the complexity of analytical aerodynamics \citep{dong2024drone}. These cross-domain successes motivate our application of autoregressive modeling to mobile robot power prediction. Recent advances in transfer learning for time-series prediction demonstrate that models trained on diverse temporal data can generalize zero-shot to unseen domains \citep{kamalov2024transfer}, suggesting that autoregressive predictors may transfer across robot instances without retraining.

\section{Data Collection}

We collected power telemetry from a single GTernal robot executing scripted motion sequences on the Robotarium testbed. The GTernal platform \citep{kim2026gternal} is a differential-drive robot with an 11 cm $\times$ 9.5 cm footprint and maximum linear speed of approximately 26 cm/s. The robot is equipped with an INA260 power monitor capable of measuring instantaneous power consumption with 10 mW precision. The onboard Raspberry Pi communicates velocity commands to a Teensy 4.0 microcontroller at 30 Hz, which executes a 100 Hz proportional-integral-derivative (PID) loop for motor control. The power monitor is polled at 10 Hz by the microcontroller, with the most recent reading returned alongside each velocity command, yielding synchronized telemetry at the 30 Hz command rate. Access to real-time power telemetry was enabled through API extensions implemented by the Robotarium team specifically to support this research.

To ensure the predictor generalizes across motion patterns rather than memorizing specific trajectories, we designed six complementary trials covering the operational envelope of the platform.

\begin{itemize}
    \item \textbf{Structured Motion.} Eight phases cycling through idle, forward-backward translation, lateral movement, circular arcs, sinusoidal paths, velocity ramps, step changes, and figure-eight patterns.
    \item \textbf{Goal Navigation.} Point-to-point navigation with proportional control to randomly sampled goal positions.
    \item \textbf{Acceleration Variation.} Controlled acceleration and deceleration profiles at multiple rates.
    \item \textbf{Extremes.} Maximum and minimum velocities including rapid direction reversals.
    \item \textbf{Random Walk.} Uniformly sampled velocity commands with smooth interpolation.
    \item \textbf{Steady State.} Extended constant-velocity segments to capture thermal equilibrium.
\end{itemize}

Each trial collected 8,000 samples over approximately 4.4 minutes, yielding a combined dataset of 48,000 samples spanning 26.5 minutes of continuous operation. Power consumption ranged from 3,000 to 8,450 mW with mean 3,652 mW and standard deviation 375 mW. Linear velocity commands spanned $-0.15$ to $0.16$ m/s while angular velocity reached $\pm 3.14$ rad/s.

\section{Power Consumption Analysis}

\begin{figure}[H]
    \centering
    \includegraphics[width=0.8\linewidth]{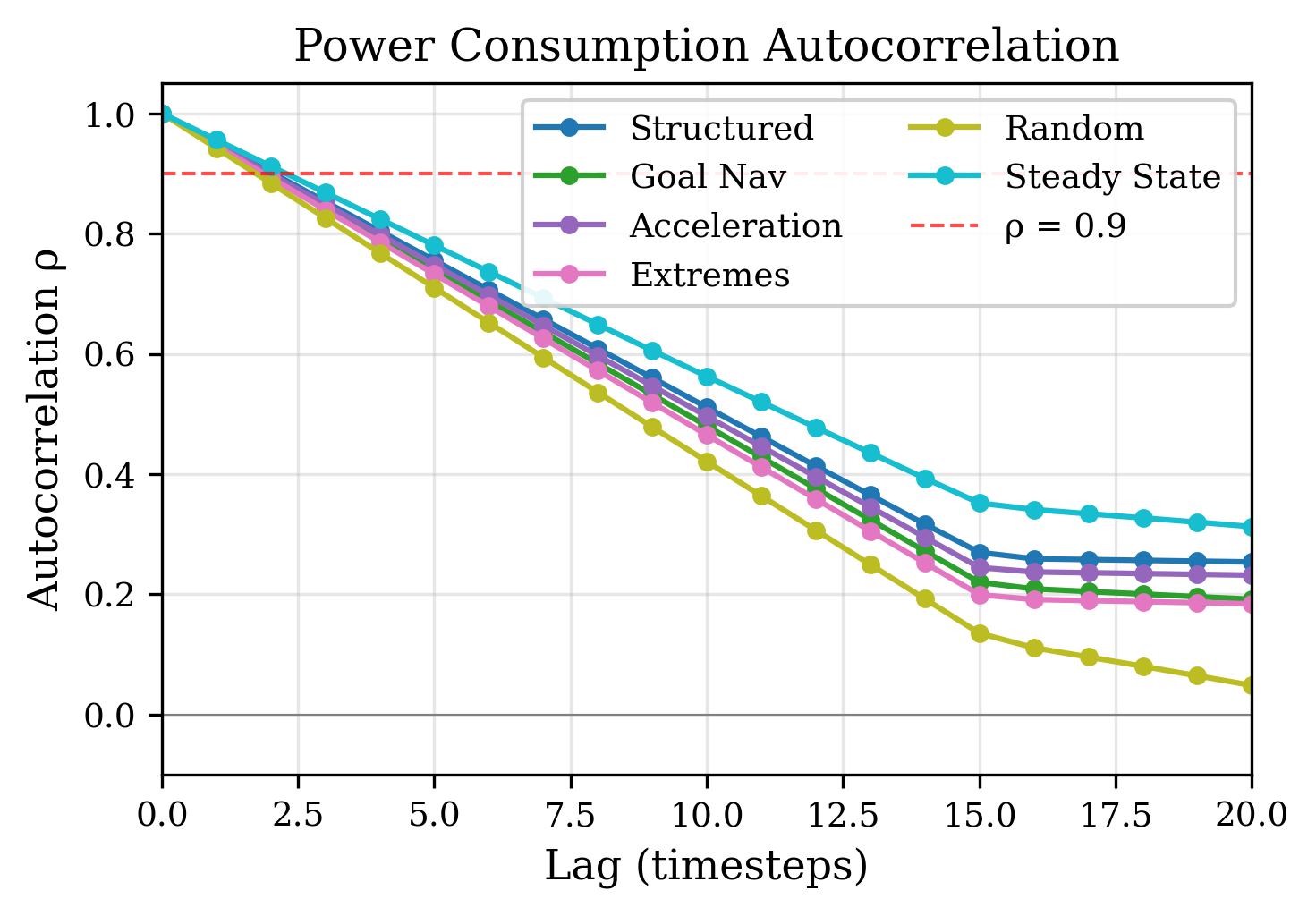}
    \caption{Power consumption autocorrelation across six motion trials. All trials exhibit lag-1 correlation $\rho_1 > 0.94$, with the structure persisting across diverse motion patterns. This strong temporal dependence motivates autoregressive prediction.}
    \label{fig:autocorrelation}
\end{figure}

Before designing a predictor architecture, we analyzed the statistical structure of power consumption to identify the most informative features. \Cref{fig:autocorrelation} shows the autocorrelation function of power consumption across all six trials. The lag-1 autocorrelation $\rho_1$ ranges from 0.942 to 0.956 with mean 0.949, indicating that consecutive power readings are highly correlated regardless of the underlying motion pattern.

This autocorrelation structure has immediate implications for predictor design. For an AR(1) process with lag-1 correlation $\rho_1$, the optimal linear predictor achieves $R^2 = \rho_1^2$ \citep{triebe2019arnet}. With $\rho_1 = 0.95$, this ceiling is approximately $R^2 = 0.90$. Any predictor that approaches this bound is extracting nearly all learnable signal from the data, with remaining variance attributable to measurement noise and genuinely unpredictable dynamics.

We verified this analysis through controlled experiments. A predictor using only velocity features including $v$, $\omega$, their derivatives, and absolute values achieves $R^2 = 0.16$. Adding a single lag of power history increases performance to $R^2 = 0.90$, with additional lags providing diminishing returns. This result confirms that power dynamics are dominated by temporal persistence rather than instantaneous kinematics.

\section{Model Architecture}

Based on the autocorrelation analysis, we design a lightweight autoregressive predictor that conditions on recent power history alongside velocity features. The model is a multi-layer perceptron with GELU activations and the following structure.

The input layer receives an 11-dimensional feature vector consisting of six velocity features and five power history lags. The velocity features are linear velocity $v$, angular velocity $\omega$, their time derivatives $\dot{v}$ and $\dot{\omega}$, and absolute values $|v|$ and $|\omega|$. The power features are the five most recent normalized power readings $\hat{P}_{t-1}, \ldots, \hat{P}_{t-5}$.

Three hidden layers with dimensions 64, 64, and 32 process the input through Gaussian Error Linear Unit (GELU) activations \citep{hendrycks2016gelu}. Unlike rectified linear units, GELU provides smooth gradients for negative inputs by weighting activations according to their magnitude under a Gaussian distribution, improving optimization stability for continuous regression tasks. Dropout with probability 0.1 is applied during training for regularization. The output layer produces a single scalar prediction $\hat{P}_t$ that is denormalized to milliwatts using the training set statistics.

All features are z-normalized using training set means and standard deviations. The total parameter count is 7,041, making the model suitable for deployment on resource-constrained platforms including the Robotarium's Raspberry Pi computers.

Training uses the Adam optimizer \citep{kingma2015adam} with learning rate $10^{-3}$ and batch size 256 for 100 epochs. Adam computes adaptive learning rates from estimates of gradient moments, providing robust convergence on non-stationary sensor telemetry. We apply early stopping with patience 10 based on validation loss. The training set consists of trials 1 through 4 while trial 5 serves as validation and trial 6 as held-out test.

\section{Results}

\subsection{Single-Robot Validation}

\begin{figure}[H]
    \centering
    \includegraphics[width=\linewidth]{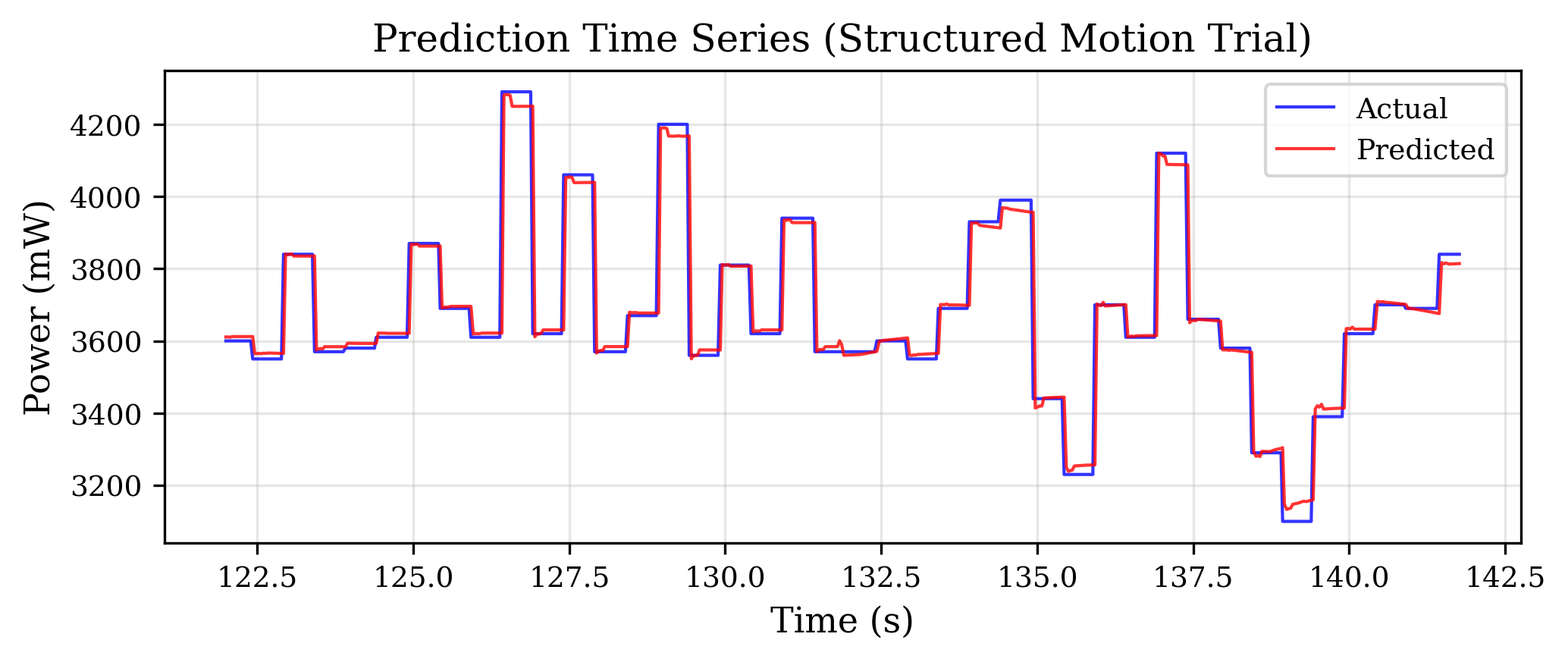}
    \caption{Time series of predicted and actual power over a 20-second segment from the structured motion trial. The predictor tracks rapid transitions and sustained levels with minimal lag.}
    \label{fig:timeseries}
\end{figure}

\Cref{fig:timeseries} shows a representative 20-second segment from the structured motion trial, illustrating how the predictor tracks both rapid transitions between motion phases and sustained power levels during steady operation. The autoregressive structure enables the model to follow abrupt changes within one or two timesteps while avoiding the oscillatory behavior that plagues purely kinematic predictors.

\begin{figure}[t]
    \centering
    \includegraphics[width=0.75\linewidth]{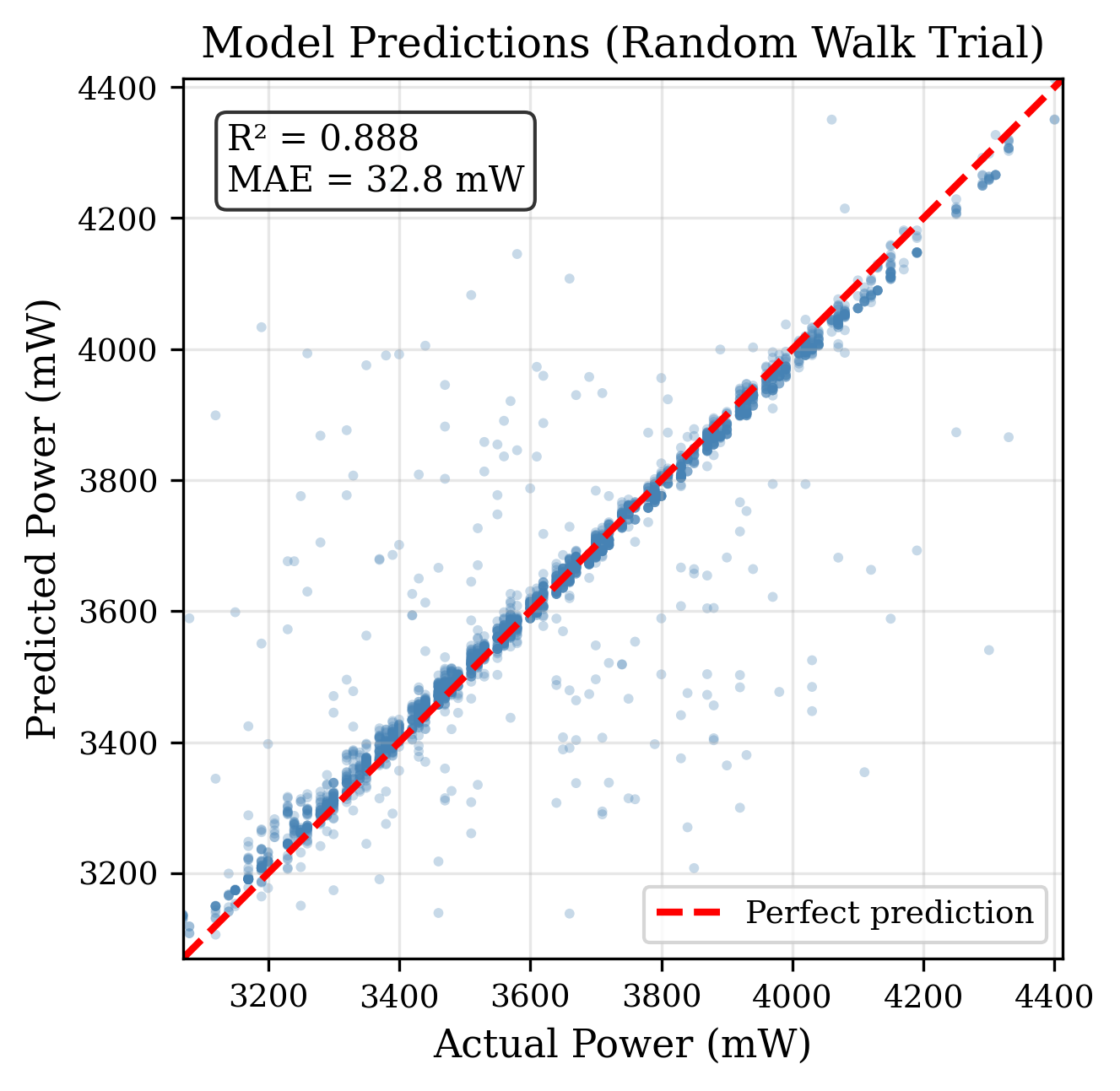}
    \caption{Predicted versus actual power consumption on the held-out steady-state trial. Points cluster tightly around the identity line with $R^2 = 0.92$ and MAE = 31.5 mW.}
    \label{fig:scatter}
\end{figure}

\Cref{fig:scatter} shows predicted versus actual power consumption on the held-out test trial consisting of steady-state motion patterns not seen during training. The predictor achieves $R^2 = 0.92$ with mean absolute error (MAE) of 31.5 mW, corresponding to 0.86\% mean absolute percentage error (MAPE) relative to the 3,652 mW mean power consumption.
Across all six trials, mean $R^2$ is 0.90 with per-trial values ranging from 0.89 to 0.92. The model performs consistently regardless of whether the motion pattern involves smooth trajectories, abrupt accelerations, or extended steady-state operation. This consistency indicates that the autoregressive structure generalizes across the operational envelope rather than fitting to specific motion primitives.

\begin{figure}[H]
    \centering
    \includegraphics[width=0.75\linewidth]{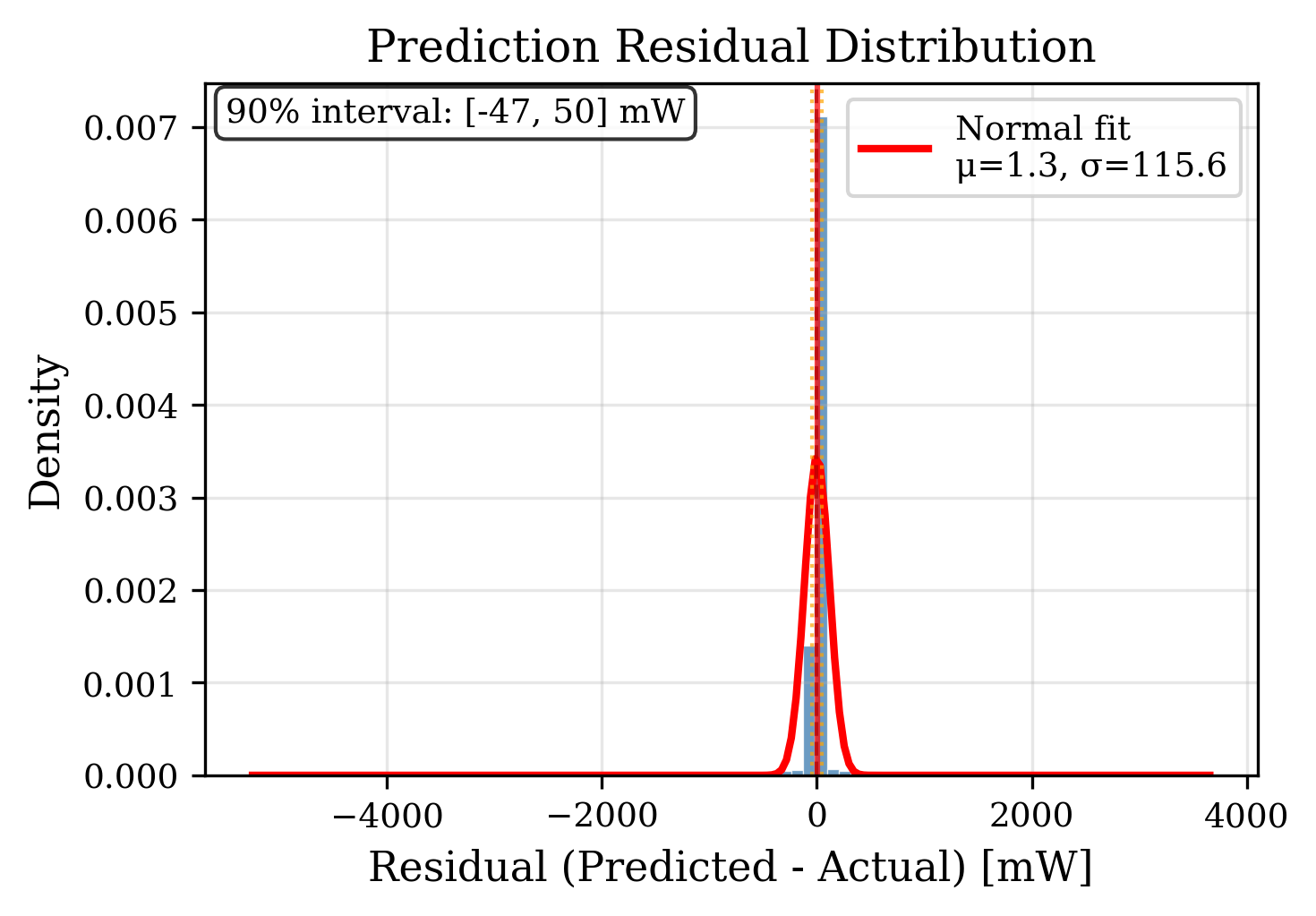}
    \caption{Distribution of prediction residuals across all trials. The distribution is approximately Gaussian with mean 1.3 mW and 90\% of errors falling within $[-47, 50]$ mW.}
    \label{fig:residual}
\end{figure}

\Cref{fig:residual} shows the distribution of prediction residuals. The mean residual is 1.3 mW, indicating the predictor is unbiased. The distribution is approximately Gaussian with 90\% of predictions falling within 50 mW of the true value. Outliers beyond 100 mW correspond to rare transient events including motor stalls and communication bursts that are intrinsically unpredictable from the available features.

\subsection{Multi-Robot Physical Validation}

To validate that the predictor transfers beyond the single robot used for training, we deployed the model on seven GTernal robots executing a random walk with collision avoidance using control barrier functions. Robot positions were tracked by the Robotarium's Vicon motion capture system at up to 120 Hz with submillimeter precision. This scenario tests generalization along two axes simultaneously. The motion pattern differs from all training trials, and the robots are physically distinct units with manufacturing variation in motors, batteries, and electronics.

\begin{figure}[H]
    \centering
    \includegraphics[width=\linewidth]{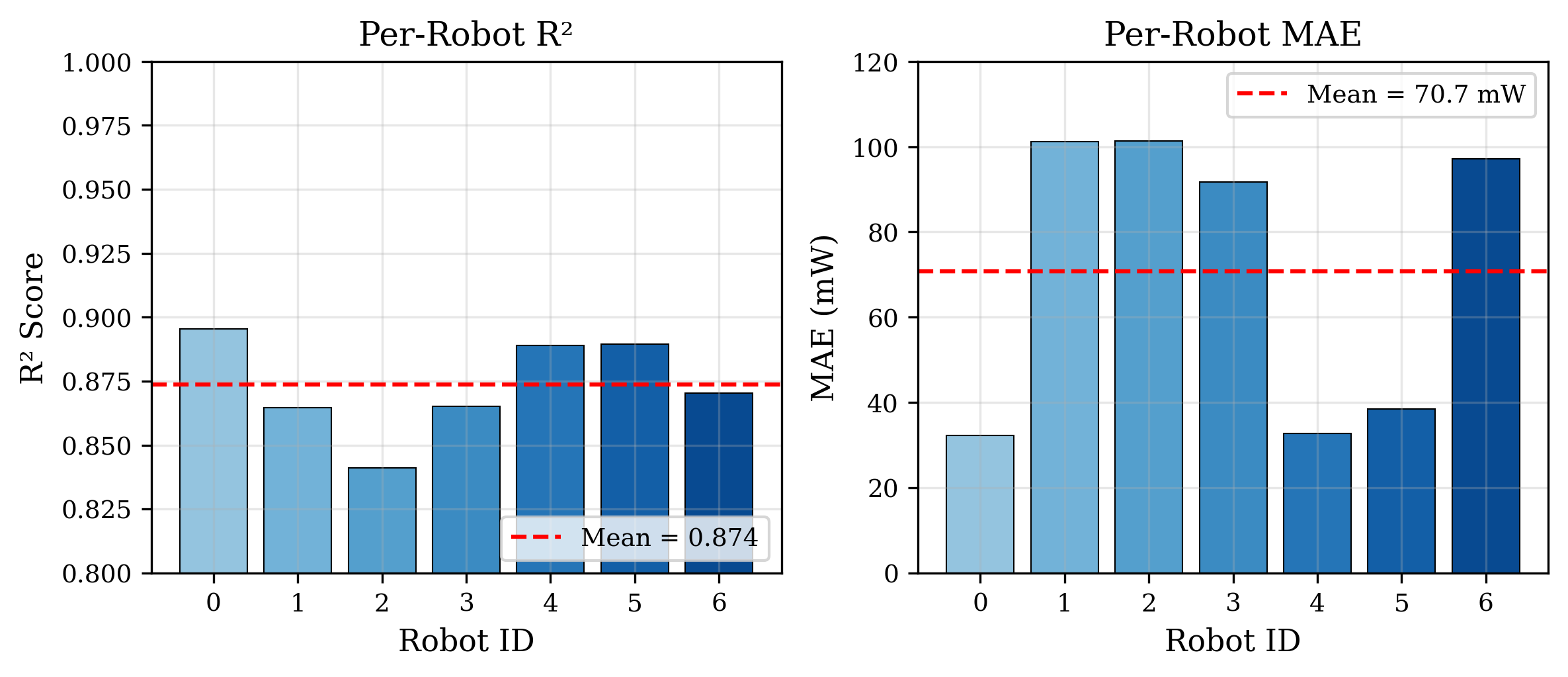}
    \caption{Per-robot $R^2$ (left) and MAE (right) during multi-robot collision avoidance. The predictor achieves mean $R^2 = 0.87$ across seven robots despite training on a single unit.}
    \label{fig:multirobot}
\end{figure}

\Cref{fig:multirobot} shows per-robot results. The predictor achieves mean $R^2 = 0.87$ with standard deviation 0.018, indicating consistent performance across the robot fleet. MAE averages 70.7 mW with variation primarily attributable to robots 1, 2, 3, and 6 which exhibit systematically higher power consumption than the training robot. Despite this distribution shift, the autoregressive structure remains effective because the temporal dynamics of power consumption are similar even when absolute levels differ.

The cumulative energy prediction error over each trial remains below 2.3\% for all robots, with four robots achieving errors under 0.1\%. This accuracy is sufficient for battery state-of-charge estimation and mission planning applications where total energy expenditure matters more than instantaneous power accuracy.

\section{Deployment}

The trained model is available in two formats for different deployment scenarios. A PyTorch checkpoint enables GPU-accelerated inference for simulation environments while a NumPy-compatible format supports deployment on Robotarium hardware without deep learning dependencies. The NumPy format runs directly on the GTernal's Raspberry Pi without requiring PyTorch installation.

For simulation, the predictor operates autoregressively by maintaining a buffer of recent predictions that serve as input for subsequent steps. This recursive structure accurately models energy accumulation over extended episodes without requiring ground-truth power readings that would be unavailable in simulation.

\begin{verbatim}
from gritsbot_energy_predictor import EnergyPredictor
predictor = EnergyPredictor('energy_predictor.npy')
predictor.reset(initial_power=3500.0)
for t in range(episode_length):
    power_mW = predictor.step(v_cmd, w_cmd)
    battery_mWh -= power_mW * dt / 3600
\end{verbatim}

For physical deployment where ground-truth power is available from the INA260, the predictor can be corrected at each timestep to prevent error accumulation.

Single-step inference completes in 224 $\mu$s on GPU, enabling throughput of 4,500 Hz. This is 150$\times$ faster than the Robotarium's 30 Hz velocity command rate, leaving ample computational budget for higher-level planning and control algorithms. The model's small parameter count of 7,041 also permits CPU inference on embedded platforms including the GTernal's Raspberry Pi at rates exceeding the control loop requirements.

\section{Conclusion}

We presented an autoregressive energy predictor for the GTernal mobile robot platform deployed in the Georgia Tech Robotarium that achieves $R^2 = 0.90$ by exploiting the strong temporal autocorrelation in power consumption dynamics. The key finding is that power history dominates kinematic features, with a single previous reading explaining 90\% of variance compared to 16\% for velocity alone. Physical validation on seven robots demonstrates zero-shot transfer to unseen units and motion patterns.

The predictor enables energy-aware multi-robot algorithms to reason about battery consumption in both simulation and physical deployment. By releasing the model and dataset, we aim to support the development of power-aware planning, task allocation, and reinforcement learning approaches for the Robotarium and similar differential-drive platforms.

\section*{Acknowledgments}

We thank Sean Wilson and Nathan Wert of the Georgia Tech Robotarium for providing access to the GTernal platform. The real-time power telemetry API used in this work was implemented by the Robotarium team to enable this research.
\bibliographystyle{unsrtnat}

\begin{thebibliography}{12}

\bibitem[Hendrycks and Gimpel(2016)]{hendrycks2016gelu}
Dan Hendrycks and Kevin Gimpel.
\newblock Gaussian error linear units ({GELU}s).
\newblock \emph{arXiv preprint arXiv:1606.08415}, 2016.

\bibitem[Kingma and Ba(2015)]{kingma2015adam}
Diederik~P. Kingma and Jimmy Ba.
\newblock Adam: A method for stochastic optimization.
\newblock In \emph{International Conference on Learning Representations (ICLR)}, 2015.

\bibitem[Triebe et al.(2019)]{triebe2019arnet}
Oskar Triebe, Nikolay Laptev, and Ram Rajagopal.
\newblock {AR-Net}: A simple auto-regressive neural network for time-series.
\newblock \emph{arXiv preprint arXiv:1911.12436}, 2019.

\bibitem[Hou et al.(2019)]{hou2019energy}
Linfei Hou, Liang Zhang, and Jongwon Kim.
\newblock Energy modeling and power measurement for mobile robots.
\newblock \emph{Energies}, 12(1):27, 2019.

\bibitem[G\'{o}ra et al.(2024)]{gora2024energy}
Krystian G\'{o}ra, Patryk Lato\'{s}, and Piotr Dudek.
\newblock Energy utilization prediction techniques for heterogeneous mobile robots: A review.
\newblock \emph{Energies}, 17(13):3256, 2024.

\bibitem[Visca et al.(2022)]{visca2022probabilistic}
Marco Visca, Sotiris Apostolopoulos, Shawn Chua, and Roland Siegwart.
\newblock Probabilistic meta-{Conv1D} driving energy prediction for mobile robots in unstructured terrains.
\newblock \emph{IEEE Access}, 10:99771--99781, 2022.

\bibitem[Ruiz et al.(2016)]{ruiz2016narx}
Luis~Gonzaga~Baca Ruiz, Manuel~Pegalajar Cuellar, Miguel~Delgado Calvo-Flores, and Mar\'{i}a~del Carmen Pegalajar~Jim\'{e}nez.
\newblock An application of non-linear autoregressive neural networks to predict energy consumption in public buildings.
\newblock \emph{Energies}, 9(9):684, 2016.

\bibitem[Dong et al.(2024)]{dong2024drone}
Shuyan Dong, Wenwu Yu, and Ahmad Alsanad.
\newblock Drone motion prediction from flight data: A nonlinear time series approach.
\newblock \emph{Systems Science \& Control Engineering}, 12(1):2409098, 2024.

\bibitem[Kamalov et al.(2024)]{kamalov2024transfer}
Firuz Kamalov, Hana Sulieman, and Sherif Moussa.
\newblock Powering electricity forecasting with transfer learning.
\newblock \emph{Energies}, 17(3):626, 2024.

\bibitem[Pickem et al.(2017)]{pickem2017robotarium}
Daniel Pickem, Paul Glotfelter, Li~Wang, Mark Mote, Aaron Ames, Eric Feron, and Magnus Egerstedt.
\newblock The {R}obotarium: A remotely accessible swarm robotics research testbed.
\newblock In \emph{IEEE International Conference on Robotics and Automation (ICRA)}, pages 1699--1706, 2017.

\bibitem[Kim et al.(2024)]{kim2026gternal}
Soobum Kim, Paden Davis, Nathan Yam, Samuel Coogan, and Sean Wilson.
\newblock {GTernal}: A robot design for the autonomous operation of a multi-robot research testbed.
\newblock In \emph{Distributed Autonomous Robotic Systems (DARS)}, pages 489--504. Springer, 2024.

\bibitem[Wilson et al.(2020)]{wilson2020robotarium}
Sean Wilson, Paul Glotfelter, Li~Wang, Siddharth Mayya, Gennaro Notomista, Mark Mote, and Magnus Egerstedt.
\newblock The {R}obotarium: Globally impactful opportunities, challenges, and lessons learned in remote-access, distributed control of multirobot systems.
\newblock \emph{IEEE Control Systems Magazine}, 40(1):26--44, 2020.

\end{thebibliography}

\end{document}